\documentclass[pdflatex,sn-mathphys-num]{sn-jnl}%

\usepackage[T1]{fontenc}%
\usepackage{ifthen}%
\usepackage{etoolbox}%
\usepackage{xargs}
\usepackage{graphicx}
\usepackage{multirow}
\usepackage{amsmath,amssymb,amsfonts}
\usepackage{amsthm}
\usepackage{mathrsfs}
\usepackage[title]{appendix}
\usepackage{xcolor}
\usepackage{textcomp}
\usepackage{manyfoot}
\usepackage{booktabs}
\usepackage{algorithm}
\usepackage{algorithmicx}
\usepackage{algpseudocode}
\usepackage{listings}
\usepackage[normalem]{ulem}%
\usepackage{float}%
\usepackage{xspace}%
\usepackage{tcolorbox}

\newcommand{\psf}{PSF\xspace}

\newcommand{\ci}{\textit{conditional imaging}\xspace}

\newcommand{\Ci}{\textit{Conditional imaging}\xspace} 

\theoremstyle{thmstyleone}

\theoremstyle{thmstyletwo}

\theoremstyle{thmstylethree}

\begin{document}

\title{Conditional Imaging for Privacy-Preserving Visual Sensing}

\author[1]{\fnm{Hengyu} \sur{Lian}}\email{lianhy0@cs.unc.edu}
\author[1]{\fnm{Debabrata} \sur{Mandal}}\email{debman@cs.unc.edu}
\author[1, 2]{\fnm{Anzhou} \sur{Wen}}\email{awen@unc.edu}
\author[1]{\fnm{Yujie} \sur{Wang}}\email{wyujie@cs.unc.edu}
\author[1]{\fnm{Xi} \sur{Peng}}\email{xipeng@cs.unc.edu}
\author*[1, 2]{\fnm{Praneeth} \sur{Chakravarthula}}\email{cpk@cs.unc.edu}
\affil[1]{\orgdiv{Department of Computer Science}, \orgname{University of North Carolina at Chapel Hill}, \city{Chapel Hill}, \postcode{27599}, \state{NC}, \country{USA}}
\affil[2]{\orgdiv{Department of Applied Physical Sciences}, \orgname{University of North Carolina at Chapel Hill}, \city{Chapel Hill}, \postcode{27599}, \state{NC}, \country{USA}}

\abstract{

Conventional visual sensing faces a fundamental dilemma: \textbf{information that should remain private today may need to be recovered tomorrow.} 
This challenge is becoming increasingly important as cameras become pervasive in our daily surroundings. 
Existing safeguards either protect visual content only after an image is formed or suppress information during capture in ways that prevent its later recovery. 
Here, we show through an information-theoretic analysis that privacy and recoverability cannot be achieved simultaneously in static optical systems.
Information preserved in a measurement for later reconstruction also remains available to an observer of that measurement, limiting the privacy such systems can provide.
To overcome this limitation, we introduce \ci, an imaging approach that changes the optical state for every captured frame. 
The measurement alone conceals the scene, while knowledge of the corresponding physical system state enables its reconstruction. 
We analyze how the optical parameters determine the information that can be recovered and use this analysis to guide the design of a camera prototype. 
We evaluate the system on a captured dataset of human faces using increasingly powerful attempts to recover the scene, from direct visualization and computational inversion to learned reconstruction and identity recognition. 
Our experiments show that \ci~preserves identity in authorized reconstructions while withholding visual information from observers without access to the system state. 
By making the physical state of the camera part of the information needed for reconstruction, \ci~enables visual information to be protected at the point of capture while remaining recoverable for future authorized use.
}
 

\newpage
\maketitle
\section{Introduction}
\label{sec1}

Cameras in personal devices, vehicles, and emerging wearables overwhelmingly record our daily surroundings \cite{gehrig2024low, loretan2025direct, chan2024detecting, lo2024ai, yun2025compact}.
Beyond serving perception in the moment, their recordings are often stored for later use, such as incident review, retrospective analysis, or tasks that are not known at the time of capture \cite{ashby2017value}.
However, these visual data can inherently capture sensitive information about individuals' identities, activities, and surroundings.
For example, a recent analysis reveals that 90\% of computer-vision papers and 86\% of downstream patents extract human data.
It further highlights that research contributing to surveillance-enabling patents has grown more than fivefold over three decades \cite{kalluri2025computer}.
Even worse, those whose privacy is affected by these recordings often have little control over how the data is captured or used \cite{zuboff2023age}.

\begin{figure}[htbp]
    \centering
    \includegraphics[width=1\textwidth]{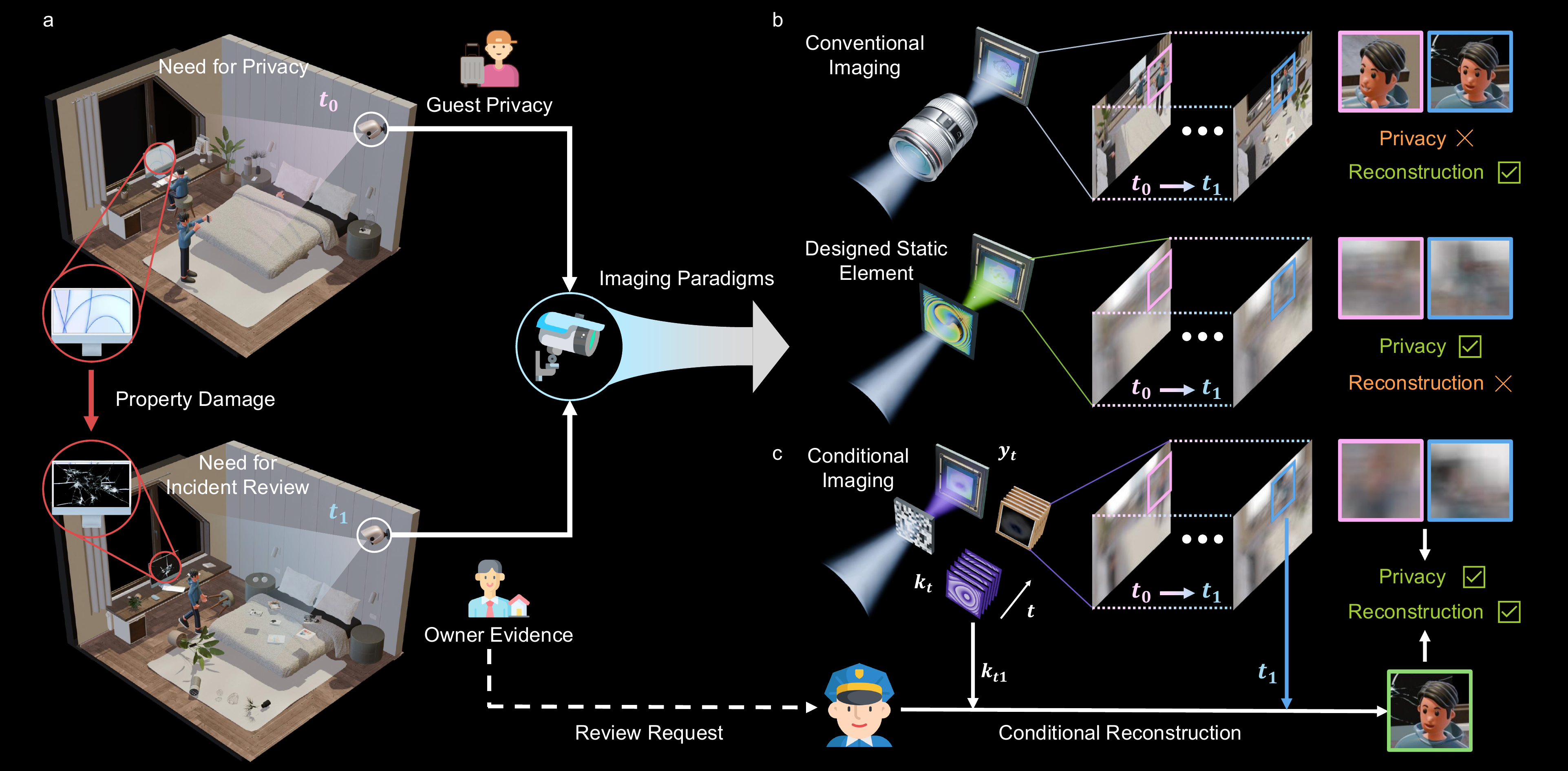}
    \caption{
    \textbf{Privacy-utility conflict and the \ci paradigm.}
    \textbf{(a)} \textit{A short-term rental scenario} illustrates the conflict.
    A host may need evidence of unexpected incidents, such as property damage, while a guest expects privacy during their stay.
    \textbf{(b)} \textit{Existing imaging paradigms}. 
    Conventional cameras form readable images before digital safeguards are applied.
    Privacy-oriented optical systems suppress sensitive information during image formation, but the suppressed information cannot be recovered later.
    \textbf{(c)} \textit{\Ci~paradigm}. 
    The optical modulation $k_t$ changes across frames, producing each measurement $y_t$ under a different imaging state.
    The measurement remains obfuscated by default, while scene recovery is conditioned on access to the corresponding state $k_t$.
    Releasing the states associated with selected frames enables authorized recovery while measurements whose states remain unavailable stay obfuscated.
    }
    \label{fig:motivation}
\end{figure}

Consider the short-term rental example illustrated in Fig.~\ref{fig:motivation}a.
A host may want an indoor camera to record evidence of unexpected events, such as property damage.
However, the same camera also captures guests during moments when privacy is expected \cite{mare2020smart, marky2020you}.
Although platforms such as Airbnb address this conflict by prohibiting indoor cameras altogether \cite{airbnb_camera_policy}, the underlying technical problem of preserving privacy while retaining the ability to recover necessary evidence remains.
Similar conflicts arise with cameras in autonomous vehicles, robots, and other systems that continuously observe their surroundings.
What is needed is a camera that keeps its recordings private by default but allows specific events to be recovered when authorized review is required.

Currently, most existing safeguards operate digitally on the recorded data through anonymization, redaction, or encryption \cite{kessler2003overview, frome2009large, ribaric2016identification}. 
Anonymization and redaction remove or modify sensitive content and are typically evaluated empirically, whereas encryption can provide stronger protection with cryptographic guarantees.
However, all of these methods act after the camera has already formed a readable image, exposing it at the sensor and during readout and processing (Fig. \ref{fig:motivation}b).

An alternative is to protect the information before a recognizable image is formed. 
Optical encodings, such as coded apertures, diffusers, and lensless systems, 
scatter scene information across the sensor \cite{javidi2025roadmap, yu2024high, li2018imaging, zhang2024memory, guo2020learning}.
While the resulting measurements may appear unrecognizable, this approach alone does not guarantee privacy. 
As the optical encoding is fixed, an observer can calibrate the system and invert the measurements or learn the inverse mapping from captured data to reconstruct the scene.
Recent privacy-oriented designs take this idea further by jointly optimizing the optics and downstream algorithms to preserve the information needed for a desired task while suppressing sensitive information, such as identity \cite{hinojosa2021learning, hinojosa2022privhar, tasneem2022learning, lopez2024privacy, cheng2024learning}.
However, these systems are usually designed for specific tasks and deployment settings.
More importantly, information suppressed by the optics is lost for everyone; even an authorized user cannot recover it later if an unforeseen need arises.

We study this problem from an information-theoretic perspective.
Our analysis reveals a fundamental limitation of static optical systems, where privacy and recoverability are inherently coupled and cannot be controlled independently.
If a measurement preserves information for later reconstruction, that information is also available to an observer who knows the imaging system.
If the information is removed for privacy, it is also lost for authorized recovery.

To break this coupling, we introduce \textit{\ci}, an imaging paradigm in which scene recovery is conditioned on the physical state of the imaging system at the time of capture (Fig.~\ref{fig:motivation}c).
The camera randomly changes its optical modulation at every frame, producing a measurement $y_t$ under a frame-specific imaging state $k_t$.
The measurement remains obfuscated by default, while access to the corresponding state $k_t$ enables state-aware reconstruction.
Thus, recoverability is conditional on access to the physical imaging state rather than being fixed by the measurement alone.
An authorized user with the corresponding state can recover the scene, whereas an observer without it must solve a state-blind inverse problem with no single fixed imaging operator to invert.
Selected frames can therefore be made recoverable by releasing their corresponding states while withholding the states associated with other frames.

We further connect this principle to the physical design of the camera.
Our analysis relates recoverable information to the spatial extent of the point spread function (\psf) and, in turn, to the physical parameters of the optical modulation, giving us a design criterion for \ci.
The criterion identifies an operating regime where the scene can be recovered with the instantaneous system state while limiting the information available without it.
We implement this criterion in a programmable spatial light modulator (SLM) prototype and evaluate it on a dataset of human faces.

Our experiments demonstrate a clear asymmetry between authorized and unauthorized access.
We evaluate identity using face verification and report the area under the receiver operating characteristic curve (AUC).
This metric measures how well a face-verification system distinguishes images of the same person from images of different people across decision thresholds.
An AUC of 0.5 corresponds to random guessing, while 1.0 corresponds to perfect separation.
With the instantaneous system state, reconstructed faces achieve an AUC of 0.91, showing that identity is well preserved for authorized verification.
Without the system state, almost all measurements do not contain enough facial structure for a face detector to locate a face.

We further stress-test privacy under a setting deliberately favorable to the unauthorized observer.
Rather than evaluating all measurements, we retain only the cases where data-driven recovery succeeds well enough for a face to be detected.
This removes the many cases where unauthorized recovery already fails at the face-detection stage.
Even under this favorable setting, the face-verification AUC drops to 0.67, compared with 0.91 for authorized reconstruction.
Thus, the reported unauthorized AUC measures identity leakage only among the most recoverable cases and not across all captured measurements.
This shows that a large gap between authorized and unauthorized access persists even when the evaluation is biased in favor of the unauthorized observer.

Together, our analysis and experiments show how privacy can be built into the image formation process.
Unlike existing optical approaches, \ci~does not achieve privacy by permanently discarding information needed for later reconstruction.
Instead, recoverability is conditioned on access to the frame-specific physical state of the imaging system.
This shifts visual privacy from deciding what information to irreversibly remove at capture to controlling which measurements can subsequently be reconstructed.

\maketitle
\section{Results}

\begin{figure}[htbp]
    \centering
    \includegraphics[width=1\textwidth]{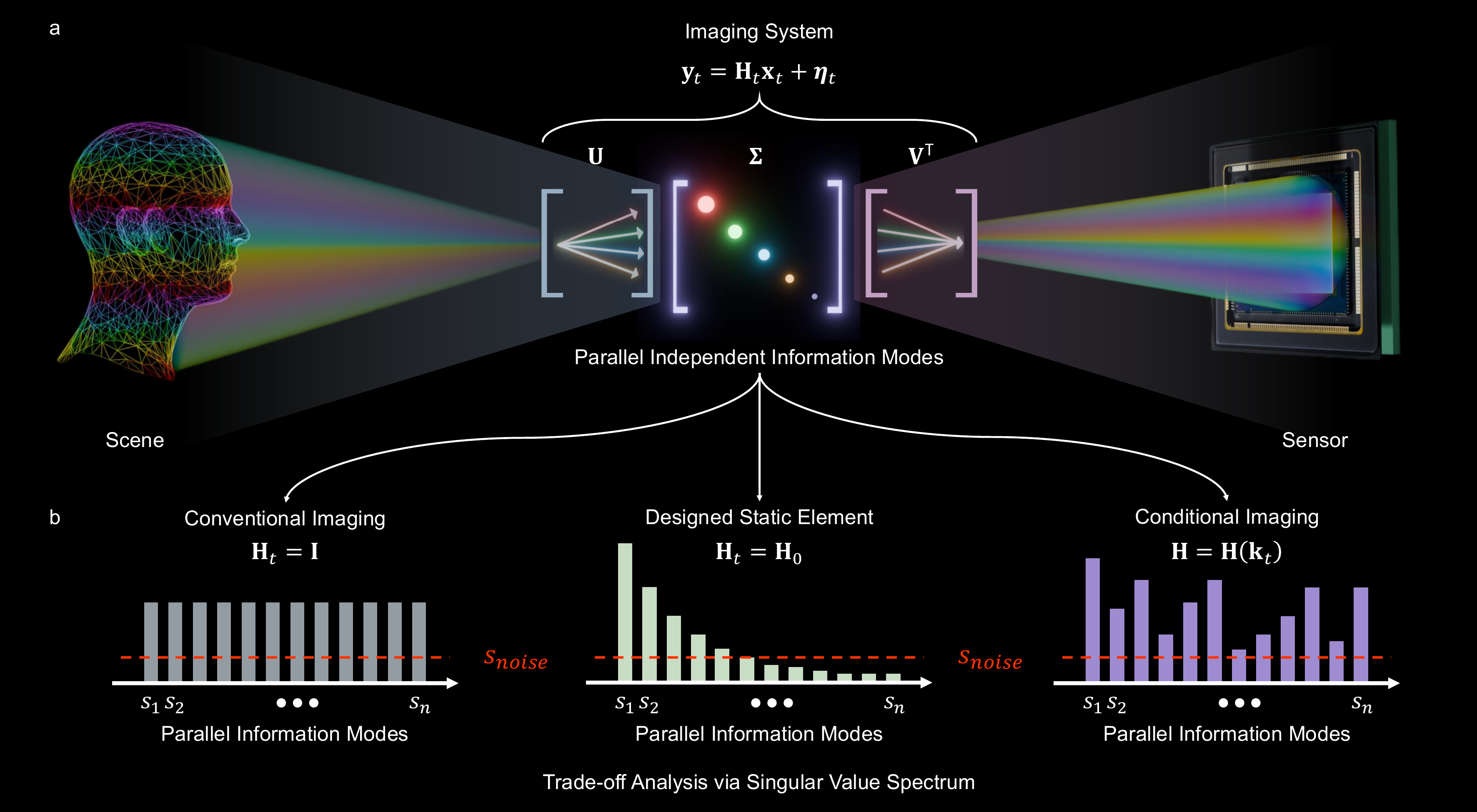}
    \caption{
    \textbf{Breaking the privacy-utility trade-off via \ci.}
    \textbf{(a) Image formation and information modes}.
    The imaging process can be mathematically modeled as a linear operation $\mathbf{y}_t=\mathbf{H}_t\mathbf{x}_t+\boldsymbol{\eta}_t$, where the system transport operator $\mathbf{H}_t$ maps a scene $\mathbf{x}_t$ to a sensor measurement $\mathbf{y}_t$ at frame $t$.
    Singular value decomposition (SVD) decomposes this mapping into a set of parallel, independent measurement channels.
    Each singular value determines how strongly information is transmitted through one channel.
    \textbf{(b) Comparison of imaging paradigms}.
    An idealized conventional imaging system ($\mathbf{H}_t=\mathbf{I}$) preserves all modes and therefore provides no privacy.
    Static optical encoding ($\mathbf{H}_t=\mathbf{H}_0$) can improve privacy by suppressing selected modes, but information removed below the noise floor is unavailable to both authorized and unauthorized observers. 
    In contrast, \ci~uses a frame-dependent operator $\mathbf{H}_t=\mathbf{H}(\mathbf{k}_t)$ that changes with the modulation state $\mathbf{k}_t$ while preserving the modes needed for stable authorized reconstruction.
    Thus, measurements can be reconstructed when $\mathbf{k}_t$ is known while remaining obfuscated when it is withheld.
    }
    \label{fig:svd}
\end{figure}

\subsection{\Ci: principle and implementation}
\label{sec:2.1}

To motivate \ci, we first consider a fundamental limitation of static optical encoding.
We model image formation as a linear mapping from a scene to a sensor measurement,

\begin{equation}
\label{eq:image_formation}
\mathbf{y}_t=\mathbf{H}_t\mathbf{x}_t+\boldsymbol{\eta}_t,
\end{equation}

where $\mathbf{x}_t$ is the scene, $\mathbf{y}_t$ is the sensor measurement, $\boldsymbol{\eta}_t$ is the noise, and $\mathbf{H}_t$ is the optical transport matrix.
For a static optical system, $\mathbf{H}_t=\mathbf{H}_0$ remains fixed across frames.

This fixed operator creates a basic limitation for privacy-preserving imaging. 
If $\mathbf{H}_0$ preserves enough information for image reconstruction, the same information is in principle available to any observer who knows or estimates $\mathbf{H}_0$.
Conversely, if the optics remove information to prevent unauthorized recovery, that information is also unavailable for later authorized reconstruction. 
Thus, a static optical operator cannot independently control privacy and recoverability.

We can formalize this coupling through the singular spectrum of the imaging operator.
As illustrated in Fig.~\ref{fig:svd}a, singular value decomposition (SVD) decomposes the transport matrix $\mathbf{H}$ into parallel measurement channels.
Each singular value $s_i$ determines how strongly the corresponding scene mode $i$ is transmitted to the sensor.
These singular values bound the mutual information $I(\mathbf{x}; \mathbf{y})$ between the scene and the measurement:

\begin{equation}
    \label{eq:mutual_information_main}
    I(\mathbf{x}; \mathbf{y}) \leq \sum_{i=1}^{R} \frac{1}{2} \log_2 \left( 1 + \frac{s_i^2 P_i}{\sigma_\text{noise}^2} \right),
\end{equation}
where $P_i$ is the signal power in mode $i$, $\sigma_\text{noise}^2$ is the noise variance, and $R$ represents the number of active modes.
Therefore, a larger singular value allows more information in the corresponding scene mode to reach the sensor.
Further details and assumptions are provided in the Supplementary Material.

Passive optics constrain how this spectrum can be shaped.
Non-negative intensity and energy conservation impose an upper bound $E_{\text{max}}$ on the total squared singular values (i.e., the energy of the singular value spectrum) of $\mathbf{H}$, $\sum_i s_i^2 \le E_{\max}$. 
Within this bound, distributing energy uniformly across all modes ($s_1 = \cdots = s_R$) maximizes information throughput. 
This represents the idealized conventional-imaging limit where the scene information is well preserved, but little optical privacy is provided (Fig.~\ref{fig:svd}b).

At the opposite extreme, minimizing information recorded while retaining a minimum operational singular spectrum energy $E_{\mathrm{req}}>0$ concentrates the remaining energy into a single mode: $s_k^2 = E_{\text{req}}, \textrm{and} \; s_{i\neq k} = 0$.
The remaining modes lie in the null space and are therefore lost to every observer (Fig.~\ref{fig:svd}b).
Task-specific optical designs \cite{hinojosa2021learning, hinojosa2022privhar, tasneem2022learning, lopez2024privacy, cheng2024learning} can instead suppress only sensitive modes, such as those carrying identity.
However, once those modes are removed by a static optic, they cannot be recovered later, even by an authorized user.
Thus, static optical encoding cannot independently control information suppression and recoverability.

\Ci changes this premise.
Instead of protecting information by permanently removing it, we make the imaging operator itself a frame-dependent variable.
In our implementation, we achieve this by placing a programmable phase modulator at the aperture of the camera.
The modulator changes the phase of light passing through different positions across the aperture, thereby changing the imaging operator for every frame:

\begin{equation}
    \label{eq:conditional_model}
    \mathbf{y}_t = \mathbf{H}(\mathbf{k}_t)\,\mathbf{x}_t + \boldsymbol{\eta}_t,
    \qquad
    \Phi_t(\mathbf{u}) = \sum_{j=1}^{K} c_j(t)\, Z_j(\mathbf{u}),
    \quad
    c_j(t) \sim \mathcal{U}[a_j, b_j].
\end{equation}
Here, $\mathbf{u}$ denotes the 2D position across the camera aperture, and $\Phi_t(\mathbf{u})$ is the phase modulation applied at that position.
We represent this phase pattern using $K$ smooth polynomial basis functions $Z_j(\mathbf{u})$.
The coefficient $c_j(t)$ controls the strength of the $j$-th basis function at frame $t$ and is randomly sampled from a bounded interval $[a_j,b_j]$.

The collection of these coefficients, $\mathbf{k}_t = (c_1(t), \dots, c_K(t))^{\mathsf{T}}$, defines the modulation state for frame $t$ and therefore determines the frame-specific imaging operator $\mathbf H(\mathbf k_t)$.
We use smooth basis functions and bounded coefficients to avoid rapid or extreme phase variations that could create deep spectral nulls and make authorized reconstruction unstable.

This construction creates an asymmetry between authorized and unauthorized recovery.
When $\mathbf{k}_t$ is known, the corresponding operator $\mathbf{H}(\mathbf{k}_t)$ is known, and scene recovery reduces to a conventional inverse problem.
When $\mathbf{k}_t$ is withheld, both the scene and the frame-specific operator are unknown, making recovery a substantially more underdetermined nonlinear inverse problem.

The remaining design question is how strongly each frame should be modulated.
Spatial phase modulation broadens the point spread function (PSF), causing light from each scene point to spread across neighboring sensor pixels.
A wider PSF better obscures recognizable structure in the raw measurement, but excessive broadening suppresses information needed for stable authorized reconstruction.
This trade-off defines a feasible range for the PSF width:
\begin{equation}
    \label{eq:design_window_main}
    \underbrace{\gamma\, d_{\text{feature}}}_{\substack{\text{privacy}}} \;\le\; \underbrace{w_{\text{PSF}}}_{\substack{\text{modulation}}} \;\le\; \underbrace{\sqrt{\Lambda / \bar{I}_{\text{req}}}}_{\substack{\text{reconstruction}}}.
\end{equation}
Here, $d_{\text{feature}}$ specifies the application-dependent feature scale to obscure, while $\bar{I}_{\text{req}}$ specifies the information required for reconstruction. 
The factors $\gamma$ and $\Lambda$ convert these requirements into PSF-width bounds based on the perceptibility threshold, PSF shape, sensor area, and SNR.
Eq.~(\ref{eq:design_window_main}) therefore defines the allowable modulation range for a given application and imaging system.
Detailed expressions and their connection to the physical modulation parameters are provided in the Supplementary Material.

\begin{figure}[htbp]
    \centering
    \includegraphics[width=1\textwidth]{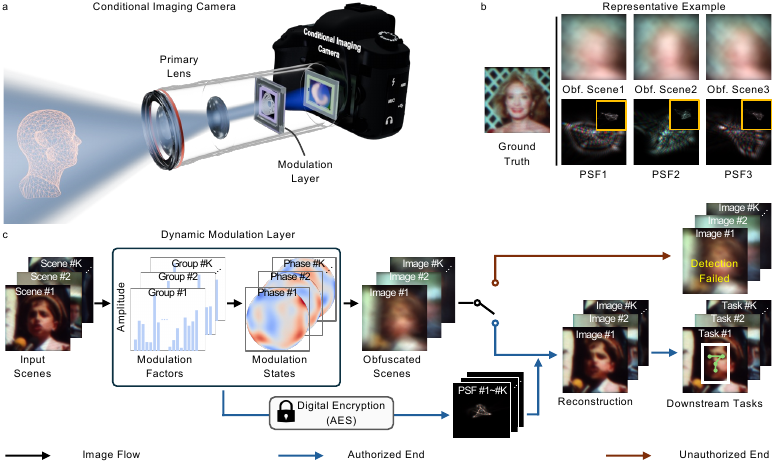}
    \caption{ 
    \textbf{\Ci~framework and operational workflow.}
    \textbf{(a)} Conceptual overview of the prototype \ci~camera, where physical image formation is governed by a time-varying optical modulation state.
    \textbf{(b)} Representative modulation examples for a single ground truth image. 
    Altering the physical state changes the frame-specific point spread function (\psf), producing completely distinct, obfuscated measurements for the exact same input scene.
    \textbf{(c)} End-to-end workflow demonstrating conditional interpretability.
    Input scenes are transformed by the dynamic modulation layer to produce noise-like, uninterpretable obfuscated measurements.
    Without the modulation state (unauthorized end), the frame-specific forward operator is unknown.
    When the modulation state is available (authorized end), the corresponding \psf~is computed and used for scene reconstruction and downstream tasks. 
    Face images from LFW public dataset \cite{huang2008labeled}.   }
    \label{fig:pipeline}
\end{figure}

Guided by this design criterion, we build a programmable optical prototype (Fig.~\ref{fig:pipeline}a).
At each frame, a spatial light modulator applies the phase pattern specified by Eq.~(\ref{eq:conditional_model}).
The full optical configuration is described in Methods and the Supplementary Material.
As shown in Fig.~\ref{fig:pipeline}b, different modulation states produce different PSFs and therefore different obfuscated measurements of the same scene.
Fig.~\ref{fig:pipeline}c summarizes the operational workflow.
The camera records an obfuscated measurement, while the corresponding modulation state is encrypted and withheld by default.
For authorized review, the state is released and used to compute the frame-specific PSF for reconstruction.
Without this state, the observer does not know the frame-specific forward operator.

We next evaluate both sides of this asymmetry. We first test whether increasingly strong computational methods can recover sensitive information without the modulation state (Sec.~\ref{privacy evaluation}). We then evaluate reconstruction quality and downstream utility when the correct state is available (Sec.~\ref{downstream tasks}).

\subsection{\Ci: resistance to unauthorized reconstruction}
\label{privacy evaluation}

\begin{figure}[htbp]
    \centering
    \includegraphics[width=1\textwidth]{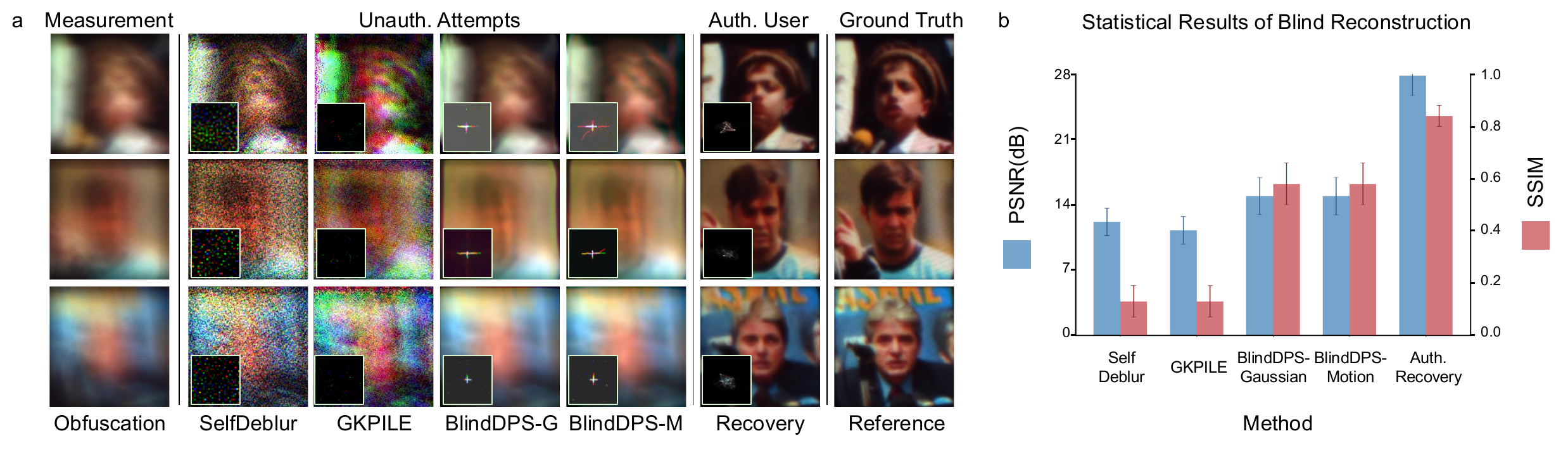}
    \caption{
    \textbf{Evaluation of blind reconstruction attempts.}
    \textbf{(a)} 
    Four state-of-the-art blind deconvolution methods---SelfDeblur, GKPILE, BlindDPS-G (Gaussian kernel), and BlindDPS-M (motion kernel)---produce substantially degraded reconstructions compared with modulation state-aware authorized recovery.
    Insets show the estimated \psf~for each method.
    \textbf{(b)} Quantitative performance comparison across the test set. 
    Authorized recovery substantially outperforms all unauthorized attempts in both PSNR (blue, left axis) and SSIM (red, right axis). Error bars denote standard deviation across the test set. 
    }
    \label{fig:attack1}
\end{figure}

\begin{figure}[htbp!]
    \centering
    \includegraphics[width=1\textwidth]{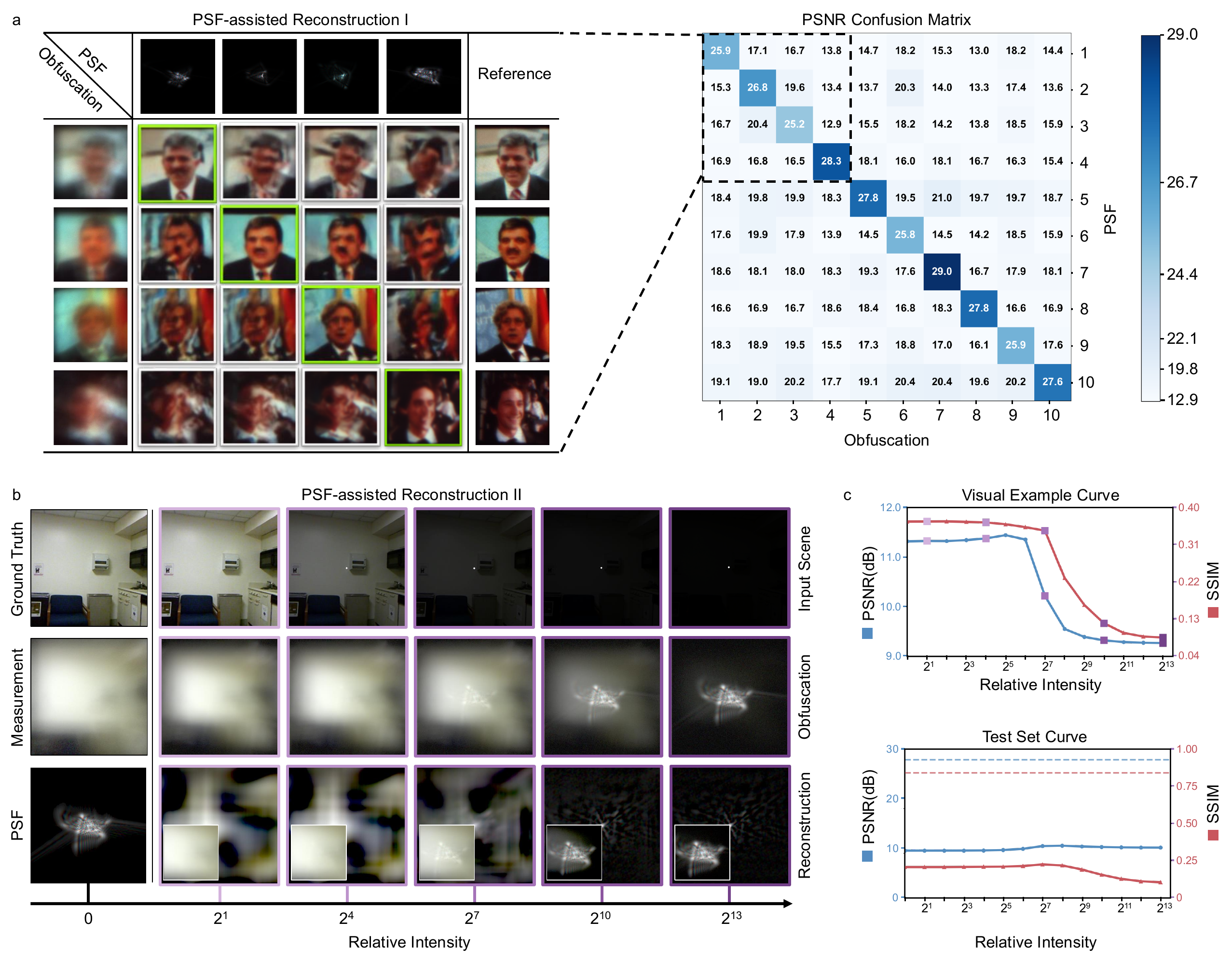}
    \caption{
    \textbf{Evaluation of \psf-assisted reconstruction attempts.}
    \textbf{(a)} \textit{Sensitivity to \psf~mismatch}. 
    The $10 \times 10$ PSNR matrix (right) shows cross-deconvolution between 10 randomly sampled measurements and \psf{}s from different modulation states.
    Correctly matched measurement--\psf~pairs (diagonal) yield high-quality reconstructions, whereas mismatched pairs (off-diagonal) suffer substantial degradation.
    The $4 \times 4$ grid (left) visualizes a subset of these combinations, with green borders indicating correctly matched pairs.
    \textbf{(b)} \textit{In-scene \psf~estimation at representative point-source intensities.}
    At low relative intensities, background structure contaminates the estimated \psf.
    At higher intensities, the point source provides a cleaner \psf~estimate but increasingly dominates the measurement, reducing the relative contribution of the surrounding scene.
    Insets show the extracted \psf{}s at representative relative intensity ratios, and image border colors correspond to the respective points in the top curves of (c).
    \textbf{(c)} \textit{Reconstruction performance across relative point-source intensities from $2^0$ to $2^{13}$}.
    PSNR (left axis, blue) and SSIM (right axis, red) are shown for the example in (b) (top) and aggregated across 500 test images (bottom), respectively.
    Consistent with the visual degradation observed in (b), both metrics remain low across all intensity levels, falling far below the authorized recovery baseline (dashed line), confirming that in-scene features do not expose sufficient information for reliable recovery.
    }
    \label{fig:attack2}
\end{figure}

\begin{figure}[htbp]
    \centering
    \includegraphics[width=1\textwidth]{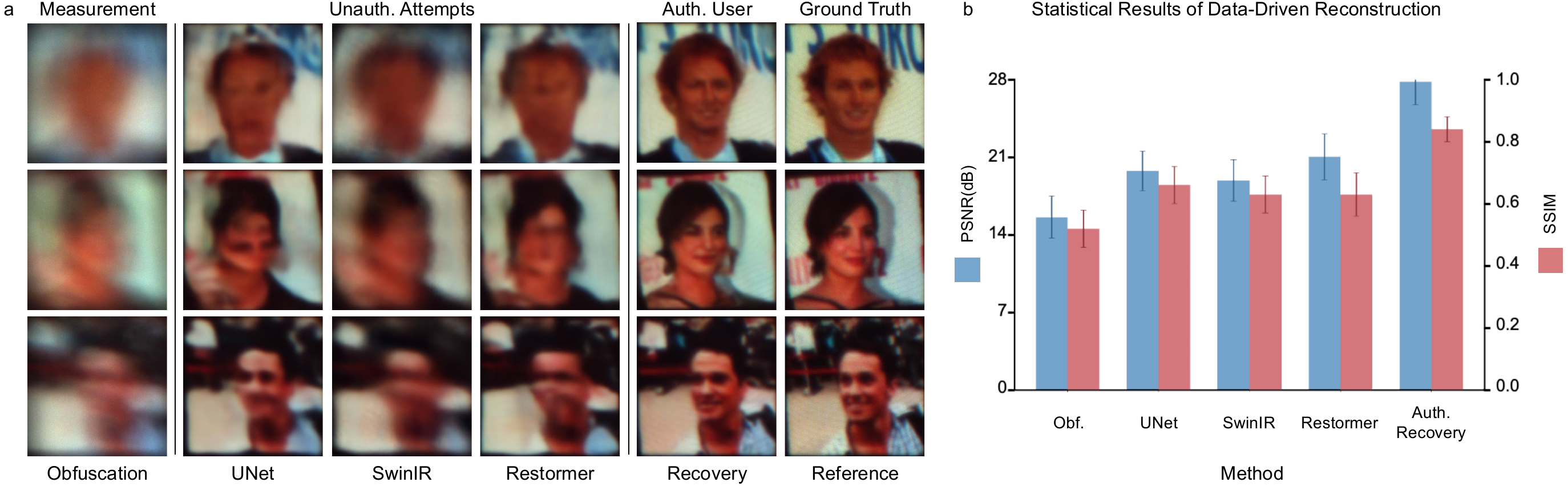}
    \caption{
    \textbf{Evaluation of data-driven reconstruction attempts.}
    \textbf{(a)} Qualitative comparison of state-blind data-driven reconstruction using UNet, SwinIR, and Restormer with state-aware authorized recovery.
    Despite training on paired obfuscated and clean images, all three models produce visibly degraded reconstructions relative to authorized recovery.
    \textbf{(b)} Quantitative comparison across the LFW test set. 
    State-aware authorized recovery substantially outperforms all data-driven reconstruction approaches and preceding unauthorized attacks in both PSNR (left axis, blue) and SSIM (right axis, red).
    Error bars denote standard deviation.
    }
    \label{fig:attack3}
\end{figure}

We evaluate whether an observer can recover scene information without access to the frame-specific modulation state $\mathbf{k}_t$.
In all experiments, the observer has access to the recorded measurements and knowledge of the camera architecture, but \textit{does not have access to the instantaneous modulation state}.
We consider progressively stronger attacks:
(\textit{i})~\textit{Blind reconstruction}, which jointly estimates the scene and unknown PSF;
(\textit{ii})~\textit{\psf-assisted reconstruction}, which attempts to estimate the instantaneous PSF from partial physical observations;
(\textit{iii})~\textit{Data-driven reconstruction}, which learns a direct inverse mapping from paired obfuscated and clean images.
We evaluate reconstruction fidelity using peak signal-to-noise ratio (PSNR) and structural similarity index measure (SSIM). We separately test whether the recovered measurements retain identity information through face detection and verification.
All experiments are conducted on the Labeled Faces in the Wild (LFW) dataset \cite{huang2008labeled}, and both quantitative and qualitative results are reported in Fig.~\ref{fig:attack1}, \ref{fig:attack2} and \ref{fig:attack3}.

\paragraph{Resistance to blind reconstruction}

We first consider blind reconstruction, where the observer has no access to the frame-specific modulation state or \psf.
The observer must therefore jointly estimate the unknown \psf and recover the scene, making reconstruction an ill-posed inverse problem.
We evaluate representative state-of-the-art methods spanning different image priors: SelfDeblur (deep image prior), GKPILE (generative prior), and BlindDPS (diffusion prior) with Gaussian (BlindDPS-G) and motion (BlindDPS-M) kernel models~\cite{ren2020neural, zhang2024blind, chung2023parallel}.
To account for slight \psf~variations across RGB channels, each method is applied independently to each channel, and the outputs are recombined.

We evaluate all methods on 2,578 face images captured with our prototype using PSNR and SSIM.
As shown in Fig.~\ref{fig:attack1}a, blind reconstruction remains low fidelity across all tested methods and fails to preserve identity-relevant facial features.
SelfDeblur and GKPILE achieve only $11.3$--$12.2$ dB PSNR and $\approx0.13$ SSIM, while the diffusion-based BlindDPS variants reach $\approx14.9$ dB PSNR and $0.58$ SSIM (Fig.~\ref{fig:attack1}b).
In comparison, authorized reconstruction using the correct modulation state achieves $27.80\pm2.06$ dB PSNR and $0.84\pm0.04$ SSIM.
These results show that the blind reconstruction methods remain substantially below the fidelity obtained when the frame-specific modulation state is known.

\paragraph{Resistance to \psf-assisted reconstruction}

Beyond completely blind reconstruction, a more informed observer might use physical knowledge of the imaging system to estimate the frame-specific \psf. 
We therefore consider two questions: \textit{how sensitive is reconstruction to \psf mismatch, and can the instantaneous \psf be estimated from features within the captured scene.}

We first quantify sensitivity to \psf mismatch by cross-deconvolving 10 randomly selected measurements with \psf{}s from 10 independently sampled modulation states.
This produces 100 measurement--\psf combinations organized in a $10 \times 10$ matrix (Fig.~\ref{fig:attack2}a).
Correctly matched pairs (diagonal entries) yield high-quality reconstructions of 25.2--29.0 dB PSNR, whereas mismatched pairs (off-diagonal entries) suffer substantial degradation.
The same behavior is visible in the qualitative examples (left $4 \times 4$ sub-matrix in Fig.~\ref{fig:attack2}a), where only correctly matched pairs produce interpretable reconstructions.
These results show that successful inversion requires an accurate estimate of the frame-specific \psf; a \psf from another modulation state is insufficient.

We next ask whether the instantaneous \psf can be estimated directly from the captured scene.
A point source provides a direct probe of the imaging system's impulse response, so we consider an observer who exploits a bright point-like feature within the scene.
To simulate this setting, we insert a white point source at the center of images from the NYU Depth V2 dataset \cite{silberman2012indoor}.
We define its relative intensity as the ratio of the point-source intensity to the mean background intensity and vary this ratio over 14 levels from $2^0$ to $2^{13}$.
The observer extracts the region surrounding the point source, subtracts the estimated background, and uses the resulting \psf~estimate for reconstruction (see Supplementary Material).

We test this strategy on 500 randomly selected images, each captured under a different modulation state.
As shown in Fig.~\ref{fig:attack2}(b, c), reconstruction remains consistently poor across the tested intensity range.
At low relative intensities, background structure contaminates the extracted \psf. 
As the relative intensity increases, the point source provides a cleaner estimate of the \psf but increasingly dominates the measurement, reducing the relative contribution of the surrounding scene. 
Consequently, both PSNR and SSIM remain well below the authorized reconstruction baseline across the tested conditions
(dashed line in Fig.~\ref{fig:attack2}c).

Together, these experiments show that reconstruction is highly sensitive to the instantaneous \psf, while the tested calibration- and scene-based strategies do not estimate that \psf accurately enough for stable recovery.

\paragraph{Resistance to data-driven reconstruction}

Finally, we consider a stronger data-driven attack in which the observer has physical access to the camera and can collect paired obfuscated and clean images.
This allows the observer to train a model that directly learns a modulation state-agnostic inverse mapping, without explicitly estimating the frame-specific \psf.
This scenario represents a deliberately favorable attack setting. 
We evaluate three representative architectures---UNet \cite{ronneberger2015u}, SwinIR \cite{Liang_2021_ICCV}, and Restormer \cite{zamir2022restormer}---spanning convolutional networks to modern transformer-based designs.
Each model is trained on 10,304 paired examples captured with our prototype and evaluated on an unseen test set of 2,578 images.

This attack improves reconstruction over the blind and \psf-assisted strategies, but a clear gap remains relative to state-aware recovery.
As shown in Fig.~\ref{fig:attack3}, UNet and SwinIR achieve average PSNRs of $19.77$ and $18.90$ dB and SSIMs of $0.66$ and $0.63$, respectively, while Restormer reaches $21.08$ dB PSNR and $0.63$ SSIM.
In comparison, authorized reconstruction using the correct modulation state achieves $27.80\pm2.06$ dB PSNR and $0.84\pm0.04$ SSIM.

Unlike modulation state-aware reconstruction, these networks must learn a single inverse across measurements generated by many different modulation states.
As the physical imaging operator changes continuously from frame to frame, a fixed inverse network cannot capture the underlying dynamics without access to the instantaneous modulation state. 
The observed performance gap is consistent with the intended effect of \ci: varying the forward operator across frames makes state-blind inversion substantially more difficult than reconstruction with the correct instantaneous state.
We next evaluate whether the residual information in these reconstructions is sufficient for face detection and identity verification.

\begin{figure}[H]
    \centering
    \includegraphics[width=1\textwidth]{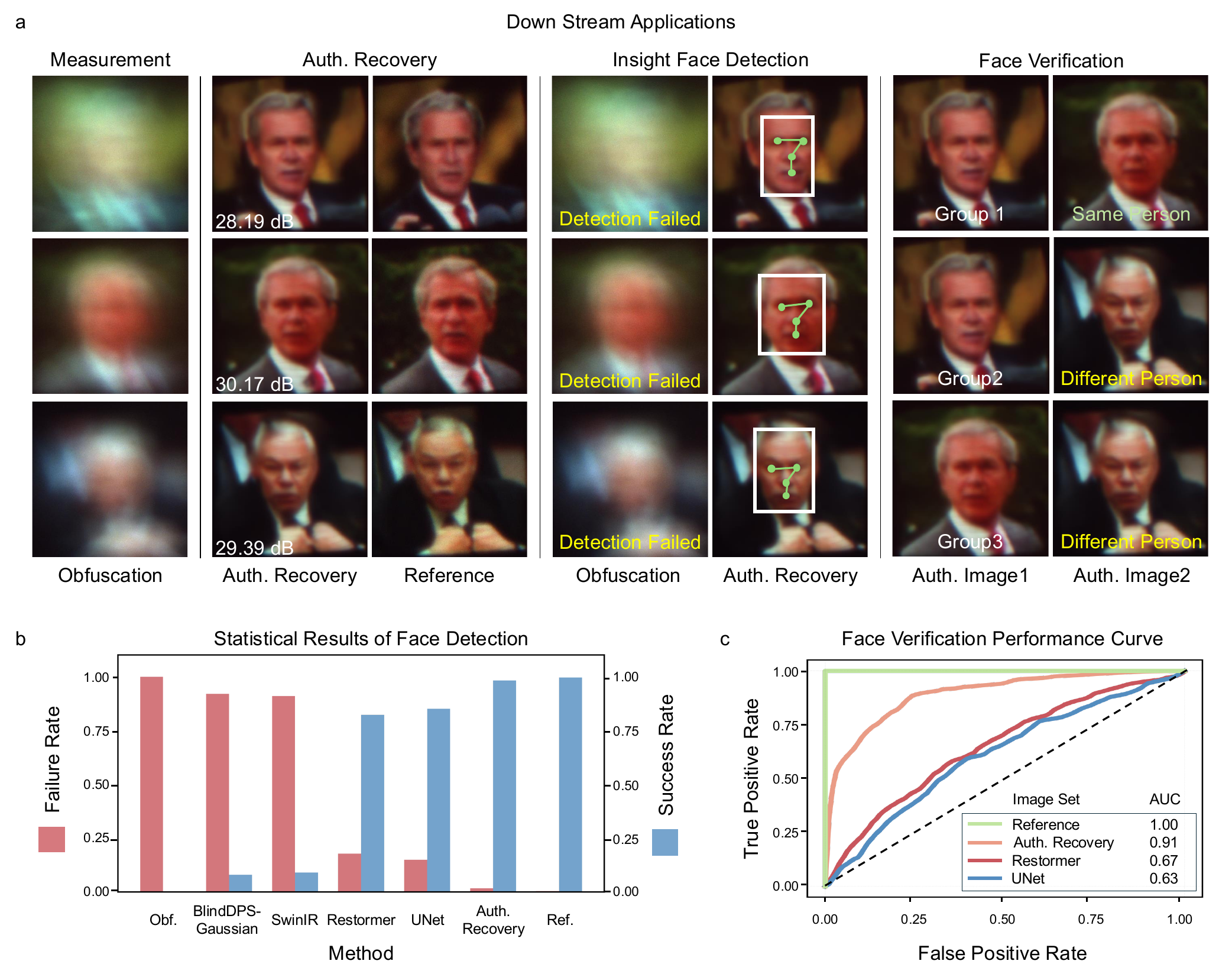}
    \caption{
    \textbf{Authorized recovery and downstream face analysis.}
    \textbf{(a)} \textit{Representative authorized reconstructions from face data captures.}
    State-aware authorized recovery restores fine spatial structure and supports downstream face analysis; PSNR values are shown for each example.
    Raw obfuscated measurements, on the other hand, result in total detection failure.
    \textbf{(b)} \textit{Face-detection success and failure rates} 
    We evaluate face-detection performance across raw measurements, unauthorized reconstructions, authorized reconstructions, and clean references.
    Raw measurements fail entirely, whereas unauthorized blind and data-driven recovery methods achieve partial success for a subset of images. 
    Authorized reconstructions achieve near-perfect success rate comparable to the clean reference.
    \textbf{(c)} \textit{Face-verification Receiver Operating Characteristic (ROC) curves}.
    For unauthorized methods, verification is evaluated only on the subset of reconstructions that \textit{successfully pass face detection}, providing a favorable conditional evaluation for the attacker.
    Authorized reconstruction achieves an area under the curve (AUC) of $0.91$, approaching the clean-reference AUC of $1.00$, while the best unauthorized recovery methods, Restormer and UNet, achieve AUCs of only $0.67$ and $0.63$, respectively.
    This shows that even among unauthorized reconstructions that successfully pass face detection, identity discrimination remains substantially degraded, whereas authorized state-aware recovery approaches clean-reference verification performance.
    }
    \label{fig:app}
\end{figure}

\subsection{\Ci: authorized recovery and downstream utility} \label{downstream tasks}

Having evaluated reconstruction without the modulation state, we now consider authorized recovery when the frame-specific state is available.
The corresponding \psf can then be computed, converting recovery from a blind to a state-aware conditional inverse problem.
Using this \psf with our authorized reconstruction method (see Supplementary Material), we recover fine spatial structure that is strongly obscured in the raw measurements (Fig.~\ref{fig:app}a).

We first evaluate whether the recovered images retain the information needed for downstream face analysis.
We quantify this performance using the detection success rate, defined as the fraction of images for which the detector successfully yields valid landmark predictions. 
As shown in Fig.~\ref{fig:app}b, raw obfuscated measurements fail face detection, whereas authorized reconstructions achieve detection rates comparable to the clean reference.
Unauthorized reconstructions from BlindDPS, SwinIR, UNet, and Restormer methods recover partial structure for detection only for a subset of images.

Face detection alone, however, does not establish whether identity is preserved.
We therefore evaluate face verification using InsightFace~\cite{deng2019arcface}.
For the unauthorized methods, we deliberately restrict this analysis to images for which \textit{face detection succeeds}, excluding cases that already fail at the detection stage.
This provides a favorable evaluation for the attacker and tests whether these successful unauthorized reconstructions retain identity information.
We select 20 individuals with five images each and evaluate 200 verification pairs per fold over 10 folds.
We report receiver operating characteristic (ROC) curves and summarize identity discrimination using area under the curve (AUC), where $\text{AUC}=1.0$ indicates perfect discrimination and $\text{AUC}=0.5$ corresponds to a chance-level success.

Authorized reconstruction achieves an AUC of $0.91$, compared with $1.00$ for the clean reference (Fig.~\ref{fig:app}c).
Under the same evaluation, Restormer and UNet achieve AUCs of $0.67$ and $0.63$, respectively.
Thus, even among \textit{unauthorized reconstructions that successfully pass face detection}, identity discrimination remains substantially degraded relative to state-aware conditional recovery.

Obfuscation need not eliminate all scene information.
We show in the Supplementary Material how coarse macro-level vision tasks, such as person counting and crowd-density estimation, can be performed directly on raw measurements without reconstructing recognizable faces.
Taken together, these results demonstrate the intended asymmetry of \ci: access to the frame-specific modulation state enables high-fidelity reconstruction and downstream analysis, while state-blind measurements and reconstructions substantially limit identity recovery under the evaluated attacks.

\maketitle
\section{Discussion}\label{sec3}
This work introduces \ci as an imaging paradigm in which scene recovery is conditioned on the physical state of the imaging system at the time of capture.
Our analysis shows why static optical systems fundamentally couple information suppression and recoverability: the same fixed imaging operator determines both how much scene information reaches the measurement and how well that information can subsequently be recovered.
A static system therefore faces an inherent privacy--recoverability trade-off: preserving information for recovery also makes it available to others, while suppressing it for privacy reduces what can later be recovered.

\Ci~introduces an additional degree of freedom by varying the imaging operator across frames and conditioning recovery on the corresponding frame-specific state.
Across our experiments, this produces a consistent operational asymmetry where state-blind unauthorized reconstruction remains substantially degraded, whereas access to the correct physical state enables high-fidelity recovery and downstream analysis.
The central advantage of this paradigm is that privacy does not require permanently discarding the information needed for later reconstruction.

Our analysis further provides a physical design criterion for this conditionality.
The criterion bounds the spatial extent of the \psf~from two sides.
Privacy sets the lower bound by requiring the \psf~to be broad enough to push features below the perceptual threshold.
Recoverability sets the upper bound by ensuring the \psf~stays compact enough to preserve necessary information for downstream tasks. 
Our prototype is designed within this regime and demonstrates the intended separation between obfuscated state-blind measurements and high-fidelity state-aware recovery.

Our current implementation demonstrates the principle of \ci using a programmable optical platform. 
A spatial light modulator provides flexible control over the frame-specific modulation, but the underlying framework is not tied to the specific hardware realization.
Emerging integrated photonic platforms, including lithium-niobate modulators with electrically tunable optical responses~\cite{yao2025integrated}, provide a potential path toward more compact and practical high-speed implementations.
We evaluate human faces as a representative privacy-sensitive domain, but the framework itself is not restricted to faces.
Extending it to other applications requires defining the relevant sensitive feature scale and reconstruction requirements for the sensing task.

More broadly, \ci shifts the role of optics in privacy-preserving sensing.
Rather than forming an immediately interpretable image and protecting it afterward, or irreversibly removing sensitive information during capture, \ci embeds privacy into the physical image formation process by making recoverability conditional on access to a frame-specific system state.
This perspective may be useful in distributed sensing, wearable cameras, augmented reality, robotics, and autonomous systems, where measurements may need to remain protected by default while retaining the possibility of authorized retrospective analysis.
\Ci balances privacy and utility not by permanently destroying information, but by controlling when captured information can be recovered.
 
\maketitle
\section{Methods}\label{sec11}

\paragraph{\Ci prototype}
We implement \ci using a programmable optical prototype with a Holoeye Leto spatial light modulator (SLM) as the modulation element.
Target scenes are displayed using a standard projector and captured through the optical system, where they are modulated by the phase patterns displayed on the SLM.
All hardware components are controlled through a computer to apply a different modulation state to each frame.
Reference images are captured under the same experimental conditions with the SLM modulation disabled.

\paragraph{Optical modulation and state management}
We parameterize the SLM phase pattern using the first 36 Zernike polynomials under Noll indexing.
For each frame, the corresponding coefficients are independently sampled from bounded uniform distributions.
The bounded coefficient ranges constrain extreme phase gradients while producing a different imaging operator for each frame.
The modulation strength is selected according to the privacy--recoverability design criterion discussed in Sec.~\ref{sec:2.1}.
The complete theoretical derivation, including the information bounds and the relationship between the modulation coefficients and \psf extent, is provided in the Supplementary Material.

Each captured frame is associated with its corresponding modulation coefficients.
These coefficients are encrypted using the Advanced Encryption Standard in Galois/Counter Mode (AES-GCM) and transmitted separately to the authorized user.
For authorized recovery, the coefficients are decrypted and used to reconstruct the frame-specific phase pattern and the corresponding \psf.

\paragraph{Data acquisition}
We evaluate \ci using the Labeled Faces in the Wild (LFW) dataset \cite{huang2008labeled} as a representative privacy-sensitive imaging task.
Each source image is displayed by the projector and physically captured through the prototype under a sampled modulation state.
We collect 10,304 paired samples for training the reconstruction models and an independent set of 2,578 samples for testing.
The captured measurements are recorded as linear intensity values under automatic exposure control with a standard white-balance setting.
Additional experiments involving controlled scene geometry use images from the NYU Depth V2 dataset \cite{silberman2012indoor}.

\paragraph{Authorized reconstruction}
Given the frame-specific modulation state, we compute the corresponding \psf and obtain the forward model for that measurement.
This converts recovery from a state-blind recovery problem into a state-aware deconvolution problem.
To account for unmodeled physical variations and mismatch between the computed and experimental \psf , we use a deep-unfolding reconstruction framework~\cite{Zhang_2020_CVPR}.
Details of the reconstruction framework, \psf~model, and optimization objectives are provided in the Supplementary Material.

\paragraph{Evaluation}
We evaluate reconstruction quality using peak signal-to-noise ratio (PSNR) and structural similarity index measure (SSIM).
The unauthorized reconstruction methods described in Sec.~\ref{privacy evaluation} are evaluated on the same 2,578-image test set, with detailed configurations and reconstruction procedures provided in the Supplementary Material.

Because reconstruction quality alone does not determine whether identity information remains accessible, we separately evaluate face detection and verification using a pretrained InsightFace model~\cite{deng2019arcface} without task-specific fine-tuning.
Identity similarity is measured using cosine similarity between ArcFace embeddings of reconstructed and reference images, and verification performance is reported using receiver operating characteristic (ROC) curves and area under the curve (AUC).
For unauthorized reconstruction methods, verification is performed \textit{only on the subset of images that successfully pass face detection}, providing a favorable conditional evaluation of the identity information remaining after successful unauthorized recovery.
Further details of the verification benchmark and evaluation protocol are provided in the Supplementary Material.

All experiments use publicly available datasets under standard protocols.
No new human-subject data are collected for this study.

\paragraph{Data availability}
All data supporting the findings of this study are available from the corresponding author upon reasonable request.

\bmhead{Author contributions}
P.C. conceived the idea and supervised the project.
H.L. and P.C. developed the approach.
H.L., A.W., and P.C. performed the theoretical analysis.
H.L. and X.P. built the prototype, collected the data, and performed the experiments.
D.M. and H.L. implemented the reconstruction methods.
A.W., H.L., Y.W., and P.C. interpreted the results and wrote the manuscript.
All authors reviewed and approved the final manuscript.

\bmhead{Conflict of interest}
The authors declare no conflict of interest.

\end{document}